\documentclass[10pt]{article}

\usepackage[T1]{fontenc}
\usepackage[utf8]{inputenc}
\usepackage{lmodern}
\usepackage{amsmath}
\usepackage{amssymb}
\usepackage{mathtools}
\usepackage[notext]{stix2}
\usepackage{booktabs}
\usepackage{array}
\usepackage{titlesec}
\usepackage[margin=1in]{geometry}
\usepackage[hidelinks]{hyperref}
\hypersetup{
  pdftitle={From the Transformer to a Formal Semantics of Coordinated Transformations},
  pdfauthor={Marco Giunti and Fabrizia Giulia Garavaglia}
}
\titlespacing*{\section}{0pt}{1.7ex plus .4ex minus .2ex}{.8ex plus .2ex}
\titlespacing*{\subsection}{0pt}{1.35ex plus .35ex minus .2ex}{.55ex plus .15ex}
\newcommand{\blank}{\text{\textemdash}}

\newcommand{\nullprop}{\mathord{\text{\textendash\,\textendash}}}

\newcommand{\wstar}{\smwhitestar}
\newcommand{\wwstar}{\smwhitestar\!\smwhitestar}

\newcommand{\starsum}{%
  \mathop{\rlap{\hspace{0.15em}\raisebox{0.15ex}{$\scriptstyle\star$}}\sum}%
}

\title{From the Transformer to a Formal Semantics of Coordinated Transformations}
\author{Marco Giunti\textsuperscript{1} and Fabrizia Giulia Garavaglia\textsuperscript{2}\\[0.45em]
\small \textsuperscript{1}ALOPHIS -- Applied Logic, Philosophy, and History of Science, University of Cagliari\\[0.2em]
\small \textsuperscript{2}Dept. of Pedagogy, Psychology, Philosophy, University of Cagliari}
\date{}

\begin{document}
\maketitle

\begin{abstract}
We develop a formal semantics for a general class of coordinated systems of parameterized transformations, taking the Transformer as a paradigmatic starting point but abstracting from its particular mathematical instantiation. Representational vectors receive both extensional reference and ordered intensional interpretations, while parameter structures acquire relative, second-level interpretations that determine how argument properties are preserved, replaced, or combined. This yields the notion of a \emph{transforming concept}: a parameterized transformation whose numerical action is coordinated with a corresponding property-level map. We then distinguish the common algebraic-semantic core of these systems from their architecture-specific functional organization, allowing different mathematical instantiations and mechanical realizations of the same abstract system type. The framework is applied to attention heads and residual connections, providing semantic explanations that depend only on the shared algebraic-semantic core and their local functional organization and therefore generalize beyond the standard Transformer. These analyses also reveal the modularity of the framework: complex architectures can be decomposed into simpler semantically coordinated subsystems whose organization may recur across different abstract system types.
\end{abstract}

\vspace{.4em}

\section{Introduction}

In previous work (Giunti and Garavaglia 2026; hereafter TR1), we proposed interpreting the Transformer as a system that transforms concepts by means of concepts. On that interpretation, the vector representations processed by the system are \emph{transformee concepts}, whereas the parameterized transformations acting on them are \emph{transforming concepts}. TR1 developed this interpretation for the Transformer architecture. The present paper asks a more general question: \emph{what formal semantics can be given to coordinated systems of transformations of this kind?}

The problem is not specific to the standard Transformer (Vaswani et al. 2017), nor even to Transformer architectures in general. We consider a broader class of systems in which matrix- or vector-parameterized transformations may operate singly or in parallel and may be organized through direct or operational output--input interconnections and output--parameter interconnections. The mathematical framework is generalized over an extended value domain and distinguishes these parameterized transformations from the non-parameterized vector-to-vector functions used in operational blocks. This yields a class broad enough to include generalized Transformer systems without being defined by Transformer-specific mathematics.

We then develop an \emph{intensional--extensional formal semantics} for these systems. Positions of representational vectors are interpreted by properties, while the vectors themselves refer extensionally to objects; the value at a position is therefore the occurrence-relative value of the property associated with that position. Parameter structures may likewise receive intensional interpretations, but they do not thereby refer to objects. Their interpretation is instead relative and second-level: it specifies how properties interpreting an argument are preserved, replaced, or combined in determining the interpretation of a result. This formal-semantic account is kept distinct from the further questions of whether and how such interpretations are representationally grounded.

The coordination between the value level and the property level yields the formal notion of a \emph{transforming concept}. A parameterized transformation is a transforming concept when its numerical action is coordinated with the property-level map determined by its relative second-level interpretation. Such transformations may be \emph{intensionally invariant} or \emph{intensionally changing} independently of whether numerical values are preserved, and dimensional preservation does not entail intensional invariance. The framework also separates the constitutive semantic organization of a transforming concept from the genesis of its parameter values and interpretations.

At a further level, we distinguish the algebraic-semantic core shared by semantically coordinated systems of transformations from their architecture-specific functional organization. This allows different mathematical instantiations and mechanical realizations of the same abstract functional architecture and clarifies the distinct routes by which abstract systems may reach the physical level. The framework is also \emph{modular}: attention heads and residual connections can themselves be treated as abstract types of semantically coordinated systems. Their semantic organization depends only on the common core and their local functional architecture, not on the overall Transformer architecture or on the particular mathematics of its standard instantiation. This makes it possible both to identify semantic invariance across mathematical variation and to distinguish semantic relations that mathematical similarity alone does not determine.

The paper proceeds as follows. Section~2 motivates the move from TR1 to the present problem and situates it within debates on representation, content, and metasemantics in AI. Section~3 develops the generalized value algebra and coordinated system architecture; Section~4 the intensional--extensional semantics; Section~5 the property algebra and transforming concepts; and Section~6 semantic coordination. Section~7 introduces abstract functional architecture, mathematical instantiation, mechanical realization, and functional-architectural invariance. Section~8 applies the framework to attention and residual connections and establishes its modularity. Section~9 discusses the philosophical significance and scope of the proposal and concludes.

\section{From the Transformer to Coordinated Systems of Transformations}

The present framework originates from the interpretation of Transformer processing developed in TR1, but its scope is broader. This section recalls the distinction between transformee and transforming concepts, situates the semantic problem within current debates on representation and content in AI, and abstracts from the Transformer to the class of coordinated systems of transformations analyzed below.

\subsection{Transformee and transforming concepts}

In TR1, we interpreted the Transformer as \emph{a system that transforms concepts by means of concepts}. The \emph{transformee concepts} are the vector representations of tokens, including both embeddings and representations generated at later stages of processing. The \emph{transforming concepts} are parameterized transformations acting on these vectors and defined by parameter matrices or vectors. Some of those parameter structures are fixed through training, whereas others are generated from the current input during inference; TR1 accordingly distinguished static from dynamic transforming concepts.

This terminology was motivated semantically as well as mathematically. A token vector was interpreted as an ordered representation whose positions correspond to properties and whose values specify the values of those properties for the represented token. Parameter matrices and vectors were instead assigned a second-level interpretation relative to the properties associated with the vector on which the corresponding transformation operates.

In the present section, however, \emph{transformee concept} and \emph{transforming concept} retain a provisional status. They identify the interpretation inherited from TR1 and the phenomenon to be generalized. A parameterized transformation does not qualify as a transforming concept merely because a matrix or vector parameterizes it. Sections~3 and 4 first specify the relevant mathematical structures and their semantics; Section~5 then distinguishes mathematical parameterization from conceptual parameterization and gives \emph{transforming concept} its formal definition.

\subsection{The semantic problem}

Recent philosophical work on AI increasingly distinguishes questions concerning concepts, representational vehicles, semantic content, misrepresentation, and representational grounding (Buckner 2018; Cappelen and Dever 2021; R\"az 2023; Freiesleben 2026; Poth and Schuster 2026; Klein 2026; Kieval 2026; Milli\`ere and Buckner 2026). These approaches differ over what makes an internal state genuinely represent a property, object, or concept. The present paper addresses a different, \emph{analytically prior} question: once relevant interpretations are specified, what formal semantic organization do the mathematical structures and transformations of such a system exhibit?

This distinction separates three questions. First, \emph{formal semantic interpretation} concerns which semantic values are assigned to mathematical structures and how those assignments are coordinated. Second, \emph{representation} concerns the conditions under which interpreted structures actually stand for objects, properties, or other contents. Third, \emph{metasemantic grounding} concerns what determines or grounds those representational relations. The present paper addresses the first question directly. In doing so, it makes structural commitments about representation: representational vectors are taken to refer to objects, their positions are interpreted by properties, and the framework specifies how such reference and property structure are preserved or transformed through processing. These commitments characterize the formal organization of representations within the framework, but they do not explain what makes a state representational, why it has one content rather than another, how such content arises, or what grounds its representational relations. In this sense, the framework is not a general theory of representation or a theory of metasemantic grounding, but a formal-semantic framework within which different such theories may be developed.

\subsection{Beyond the Transformer}

Nothing in this problem restricts the analysis to the Transformer. At the functional level, Transformer processing contains parameterized transformations of vectors, parallel organizations of such transformations, and several ways of coordinating their outputs. Outputs may become later inputs directly, may be regrouped, may pass through mathematically specified operational blocks, or may become parameters of later transformations.

TR1 described the last relation mechanically as an \emph{output--weight interconnection}, because the relevant parameters function as weights of simple neural networks. At the more general functional level adopted here, we call it an \emph{output--parameter interconnection}. The same abstraction also requires a distinction between parameterized transformations, which provide the basis for transforming concepts, and non-parameterized vector-to-vector functions used in operational blocks. Operational blocks may modify a signal and mediate output--input relations, but they are not transformations in the technical sense adopted in this paper.

The resulting coordinated systems may contain transformations and/or parallel operations organized through direct output--input, operational output--input, and output--parameter interconnections. The Transformer is an important instance of this broader organization, but the class is not defined by Transformer-specific mathematical operations or by any particular neural realization.

The next section develops the mathematical framework required to characterize this class independently of its particular instances.

\section{Generalized Transformations, the Value Algebra, and Their Coordination}

The semantic framework developed in the following sections requires a mathematical description that does not depend on the particular numerical operations used by the standard Transformer. We therefore begin by generalizing the notions of value, vector, matrix, and parameterized transformation over a blank-extended numerical domain. This yields an abstract value algebra in which ordinary matrix-defined transformations, componentwise multiplicative transformations, and vector additions can be treated within a single construction. We then introduce the forms of parallel organization and interconnection through which transformations can be coordinated into a system, distinguishing parameterized transformations from the non-parameterized vector-to-vector functions involved in operational output--input interconnections.

\subsection{Generalized values, vectors, and matrices}

Let
\[
\overline{\mathbb R}
:=
\mathbb R\cup\{\blank\},
\]
where $\blank$ is a distinguished \emph{blank} element. The blank is not a further numerical value and, in particular, is not identified with $0$. Its role is to represent the absence of a contribution within the same domain in which ordinary numerical values occur.

Let
\[
\star,\diamond:
\mathbb R\times\mathbb R
\longrightarrow
\mathbb R
\]
be arbitrary binary operations. No assumption is made that they are associative, commutative, distributive, or distinct from one another. Each operation $\omega\in\{\star,\diamond\}$ can be extended to $\overline{\mathbb R}$ in two different ways.

For each \(\omega\in\{\star,\diamond\}\), we define its additive and multiplicative extensions to \(\overline{\mathbb R}\) by, for all \(x,y\in\mathbb R\),
\[
x\mathbin{\omega^{+}}y:=x\mathbin{\omega}y,
\qquad
x\mathbin{\omega^{\times}}y:=x\mathbin{\omega}y;
\]
\[
\blank\mathbin{\omega^{+}}\blank:=\blank,
\qquad
\blank\mathbin{\omega^{\times}}\blank:=\blank;
\]
\[
\blank\mathbin{\omega^{+}}x:=x,
\qquad
x\mathbin{\omega^{+}}\blank:=x,
\qquad
\blank\mathbin{\omega^{\times}}x:=\blank,
\qquad
x\mathbin{\omega^{\times}}\blank:=\blank.
\]
Thus, the additive extension \(\omega^{+}\) makes the blank value \(\blank\) a two-sided identity element, whereas the multiplicative extension \(\omega^{\times}\) makes it a two-sided absorbing element. On pairs of real numbers, both extensions coincide with the original operation \(\omega\).

The labels \emph{additive} and \emph{multiplicative} therefore characterize the role played by the blank, not the algebraic nature of the underlying operation. The same ordinary operation can consequently occupy either role. Ordinary addition, for example, gives rise both to $+^{+}$, in which the blank is neutral, and to $+^{\times}$, in which it is absorbing.

A \emph{generalized vector} of dimension $n$ is an ordered $n$-tuple
\[
\boldsymbol x
=
(x_1,\ldots,x_n)
\in
\overline{\mathbb R}^{\,n},
\]
and a \emph{generalized matrix} is an array
\[
\boldsymbol M
=
(m_{ij})
\in
\overline{\mathbb R}^{\,m\times n}.
\]
Once the domain is clear, we will normally omit the adjective \emph{generalized}.

Since the operation occupying the additive role need not be associative, its finite iteration requires an explicit association convention. Using the generalized finite-sum symbol with the star placed in the concavity of $\Sigma$, we define
\[
\starsum_{i=1}^{1}a_i:=a_1,
\qquad
\starsum_{i=1}^{k+1}a_i
:=
\left(
\starsum_{i=1}^{k}a_i
\right)\star^{+}a_{k+1}.
\]
Thus, the iteration is left-associated.

This abstract structure defines what we will call the \emph{value algebra}. Its crucial feature is not the use of any particular pair of numerical operations, but the distinction between two abstract roles: the multiplicative operation $\diamond^{\times}$ determines the contribution made by an input value to an output position, while the additive operation $\star^{+}$ combines the contributions directed to that position.

\subsection{Parameterized vector transformations and the value algebra}

Parameterized transformations require one additional value used specifically in parameter structures. We introduce the special \emph{preserving value} \(\wstar\) and define the extended parameter-value domain
\[
\overline{\mathbb R}_{\wstar}
:=
\overline{\mathbb R}
\cup
\{\wstar\}.
\]

The operation occupying the multiplicative role is extended on its right-hand argument by stipulating that $\wstar$ is a right identity:
\[
x\diamond^{\times}\wstar
:=
x
\qquad
\text{for every }x\in\overline{\mathbb R}.
\]

The special value $\wstar$ is therefore a parameter value, not an additional possible value of ordinary argument or resulting vectors. These continue to have values in $\overline{\mathbb R}$.

Let
\[
\boldsymbol M
=
(m_{ij})
\in
\overline{\mathbb R}_{\wstar}^{\,m\times n}
\]
and
\[
\boldsymbol x
=
(x_1,\ldots,x_m)
\in
\overline{\mathbb R}^{\,m}.
\]
For each output position $j$, the matrix entry $m_{ij}$, together with the input value $x_i$, determines the contribution \(x_i\diamond^{\times}m_{ij}.\) The contributions directed to the same output position are then combined by the operation occupying the additive role. We therefore define the \emph{matrix-parameterized transformation}
\[
L_{\boldsymbol M}^{(\star^{+},\diamond^{\times})}
:
\overline{\mathbb R}^{\,m}
\longrightarrow
\overline{\mathbb R}^{\,n}
\]
by
\[
\left(
L_{\boldsymbol M}^{(\star^{+},\diamond^{\times})}
(\boldsymbol x)
\right)_j
:=
\starsum_{i=1}^{m}
\left(
x_i\diamond^{\times}m_{ij}
\right),
\qquad
j=1,\ldots,n.
\]

The construction preserves the division of labor found in ordinary vector-matrix multiplication while dispensing with its specific algebraic operations:
\[
\boxed{
\begin{array}{l}
\diamond^{\times}
\text{ determines the contribution of each input position to an output position},\\[1mm]
\star^{+}
\text{ combines the contributions directed to that output position}.
\end{array}}
\]
The construction therefore generalizes the matrix-defined linear transformations obtained by ordinary vector-matrix multiplication. Linearity can still be defined in the usual way relative to the two generalized operations that determine a matrix-defined transformation; however, in the absence of additional algebraic axioms, $L_{\boldsymbol M}^{(\star^{+},\diamond^{\times})}$ need not be linear relative to those operations.

The matrix $\boldsymbol M$ \emph{parameterizes} the transformation in a purely mathematical sense: fixing $\boldsymbol M$ selects one transformation from the corresponding family. Nothing in this definition requires its entries to have been learned, dynamically generated, or semantically interpreted. Mathematical parameterization will therefore remain distinct from the semantic interpretation of parameter structures developed in Sections~4 and~5.

A second important family consists of transformations parameterized by a vector and operating componentwise. Given
\[
\boldsymbol v
=
(v_1,\ldots,v_n)
\in
\overline{\mathbb R}_{\wstar}^{\,n},
\]
we define the \emph{vector-parameterized componentwise transformation}
\[
C_{\boldsymbol v}^{\diamond^{\times}}
:
\overline{\mathbb R}^{\,n}
\longrightarrow
\overline{\mathbb R}^{\,n}
\]
by
\[
C_{\boldsymbol v}^{\diamond^{\times}}(\boldsymbol x)
:=
\boldsymbol x\diamond^{\times}\boldsymbol v,
\]
that is,
\[
\left(
C_{\boldsymbol v}^{\diamond^{\times}}(\boldsymbol x)
\right)_i
:=
x_i\diamond^{\times}v_i.
\]

We can now define the generalized diagonal matrix associated with $\boldsymbol v$:
\[
\boldsymbol D(\boldsymbol v)
:=
\begin{pmatrix}
v_1 & \blank & \cdots & \blank\\
\blank & v_2 & \cdots & \blank\\
\vdots & \vdots & \ddots & \vdots\\
\blank & \blank & \cdots & v_n
\end{pmatrix}.
\]
The off-diagonal entries contain blanks rather than zeros because they express the absence of a contribution rather than a numerical contribution equal to zero.

\textbf{Diagonal Representation Theorem.} For every
\[
\boldsymbol v
\in
\overline{\mathbb R}_{\wstar}^{\,n},
\]
\[
\boxed{
L_{\boldsymbol D(\boldsymbol v)}^{(\star^{+},\diamond^{\times})}
=
C_{\boldsymbol v}^{\diamond^{\times}}
=
L_{\boldsymbol D(\boldsymbol v)}^{(\diamond^{+},\diamond^{\times})}.
}
\]

In the $j$-th output position, every off-diagonal entry of $\boldsymbol D(\boldsymbol v)$ is blank. Since the blank is absorbing under $\diamond^{\times}$, all corresponding contributions are blank. The only non-blank contribution is \(x_j\diamond^{\times}v_j.\) Since the blank is neutral under either $\star^{+}$ or $\diamond^{+}$, aggregating that contribution with the remaining blanks leaves it unchanged. Thus, in the diagonal case, the resulting componentwise transformation depends on the operation occupying the multiplicative role, not on which operation occupies the additive role.

When the relevant parameter values are ordinary real values and the two operational roles are instantiated by ordinary addition and multiplication, the usual matrix-defined linear transformation is recovered as
\[
\mathcal L[\boldsymbol M]
=
L_{\boldsymbol M}^{(+^{+},\times^{\times})}.
\]

For componentwise multiplication,
\[
\boxed{
\circ[\boldsymbol v]
=
L_{\boldsymbol D(\boldsymbol v)}^{(+^{+},\times^{\times})}
=
C_{\boldsymbol v}^{\times^{\times}}
=
L_{\boldsymbol D(\boldsymbol v)}^{(\times^{+},\times^{\times})}.
}
\]

For vector addition,
\[
\boxed{
+[\boldsymbol v]
=
L_{\boldsymbol D(\boldsymbol v)}^{(\times^{+},+^{\times})}
=
C_{\boldsymbol v}^{+^{\times}}
=
L_{\boldsymbol D(\boldsymbol v)}^{(+^{+},+^{\times})}.
}
\]

These equalities show that, in the diagonal case, the componentwise transformation is determined by the operation occupying the multiplicative role, while the additive role is irrelevant. Thus ordinary matrix-defined transformation \(\mathcal L[\boldsymbol M]\), componentwise multiplication \(\circ[\boldsymbol v]\), and vector addition \(+[\boldsymbol v]\) are unified within the generalized value algebra.

In what follows, unless explicitly stated otherwise, the term \emph{transformation} will be used in a technical sense to denote a parameterized transformation of one of the two kinds just defined: a matrix-parameterized transformation $L_{\boldsymbol M}^{(\star^{+},\diamond^{\times})}$ or a vector-parameterized transformation $C_{\boldsymbol v}^{\diamond^{\times}}$. Other vector-to-vector mathematical functions may occur within the systems considered below, most importantly as operational blocks, but they are not transformations in this technical sense.

\subsection{Parallel transformations}

The transformations just defined need not operate sequentially one at a time. Let
\[
T_k:
\overline{\mathbb R}^{\,m}
\longrightarrow
\overline{\mathbb R}^{\,n},
\qquad
k=1,\ldots,s,
\]
be a finite family of transformations with common input and output arities, and let
\[
\boldsymbol x_1,\ldots,\boldsymbol x_s
\in
\overline{\mathbb R}^{\,m}.
\]
The transformations may operate in parallel, producing
\[
T_1(\boldsymbol x_1),
\ldots,
T_s(\boldsymbol x_s).
\]

The transformations in the family need not be identical. What matters for the present construction is that their input vectors have the same arity $m$ and their output vectors the same arity $n$. The input vectors can therefore be arranged as the rows of a matrix
\[
\boldsymbol X
:=
\begin{pmatrix}
\boldsymbol x_1\\
\vdots\\
\boldsymbol x_s
\end{pmatrix}
\in
\overline{\mathbb R}^{\,s\times m},
\]
and the corresponding outputs as the rows of a matrix
\[
\boldsymbol Y
:=
\begin{pmatrix}
T_1(\boldsymbol x_1)\\
\vdots\\
T_s(\boldsymbol x_s)
\end{pmatrix}
\in
\overline{\mathbb R}^{\,s\times n}.
\]

A particularly important special case is obtained when \(T_1=\cdots=T_s=T,\) so that the same transformation is applied row-wise to every vector.

The more general case in which the $T_k$ differ is essential. In standard Transformer processing, positional-encoding addition, masking, and residual addition provide important cases in which different rows are subjected to transformations parameterized by different vectors. If the parameter vectors are \(\boldsymbol p_1,\ldots,\boldsymbol p_s,\) the corresponding family has the form
\[
+[\boldsymbol p_k]:
\boldsymbol x_k
\longmapsto
\boldsymbol x_k+\boldsymbol p_k,
\qquad
k=1,\ldots,s.
\]

At this stage, however, parallelism is considered only in mathematical terms, as a structural characteristic of the architecture. No semantic condition is imposed on the vectors involved. The additional requirement of \emph{interpretive uniformity} will be introduced only in Section~6.

\subsection{Direct output--input interconnections}

Transformations become parts of a coordinated system when their outputs and inputs are connected according to an architecture. An \emph{output--input interconnection} is a structural relation through which output values at one stage become available as inputs at a subsequent stage. In a \emph{direct output--input interconnection}, those values are not modified by any further mathematical function.

A \emph{standard output--input interconnection} passes an output vector to a subsequent transformation while preserving the relevant processing-stream organization; in the standard Transformer, this includes token-wise correspondence. A \emph{routing output--input interconnection} instead reorganizes or regroups unmodified outputs to form a required input structure, as in Concat. The two labels need not be exclusive: Transformer Concat also preserves token-wise correspondence.

Thus direct interconnections may transmit, regroup, or redirect values, but do not mathematically modify them.

\subsection{Operational blocks and operational output--input interconnections}

A different form of output--input coordination occurs when values produced at one stage become inputs at a subsequent stage only after being mathematically modified. An \emph{operational output--input interconnection} is the complete architectural relation in which this modification is mediated by an \emph{operational block}.

Typical operational blocks in the Transformer include the vector functions \(\operatorname{reLU}\), \(\operatorname{softmax}\), and \(\operatorname{norm}\), corresponding respectively to the matrix operations \(\operatorname{ReLU}\), \(\operatorname{Softmax}\), and \(\operatorname{Norm}\) considered in TR1.

In the simplest case, \(T_1 \longrightarrow F \longrightarrow T_2.\) Here, \(T_1\longrightarrow F\) identifies a \emph{source} from which the operational block receives its input, while \(F\longrightarrow T_2\) identifies a \emph{destination} to which its output is sent. These relations are not themselves separate operational output--input interconnections. The operational output--input interconnection is the complete mediated relation between the upstream and downstream parts of the system through $F$.

More generally, an operational block may have multiple sources and/or destinations. At the abstract architectural level it is characterized by these relations, its input--output arity $n$, and its \emph{functional role}; a particular mathematical instantiation supplies a dimension-preserving function
\[
F:\overline{\mathbb R}^{\,n}\longrightarrow\overline{\mathbb R}^{\,n},
\qquad
F(\boldsymbol x):=\bigl(f_1(\boldsymbol x),\ldots,f_n(\boldsymbol x)\bigr),
\]
that performs that role. The semantic reason for dimensional preservation will be given in \S5.3. Operational interconnections may also be \emph{interwoven} in the sense of TR1, drawing values contributing to an output across different parallel streams.

The fundamental distinction is therefore
\[
\boxed{
\begin{array}{l}
\text{direct output--input interconnection: transmitted values are not mathematically modified},\\[1mm]
\text{operational output--input interconnection: transmitted values are modified by an operational block}.
\end{array}}
\]

Operational blocks must also be distinguished from transformations in the technical sense fixed in \S3.2. An operational block is mathematically a vector-to-vector function, but it is not a transformation in the terminology adopted here, since it is not parameterized by a matrix or vector of the specified kinds.

\subsection{Output--parameter interconnections}

An output may also determine all or part of the \emph{parameter structure} of a subsequent transformation. We call this an \emph{output--parameter interconnection}. Schematically, if \(T_1(\boldsymbol x)=\boldsymbol p,\) the output may parameterize a later transformation \(T_{\boldsymbol p}(\boldsymbol y)\); analogously, output matrices or coordinated families of outputs may supply all or part of later parameter matrices or vectors. What defines the relation is that the relevant output values are used \emph{as parameters rather than as the input argument} of the subsequent transformation.

The present notion generalizes the \emph{output--weight interconnections} introduced in TR1. There, parameterized vector transformations were identified with simple neural networks, so dynamically generated parameters appeared specifically as neural-network weights. At the present functional and mathematical level, no particular mechanical realization is assumed.

We retain throughout the technical distinction fixed above: a \emph{transformation} is one of the parameterized vector-to-vector functions of \S3.2, whereas an \emph{operation} is a parallel organization of transformations with common input and output arities (\S3.3). This architectural use of \emph{operation} is distinct from the algebraic operations $\star$ and $\diamond$.

\emph{With this terminology, a coordinated system of transformations consists of transformations and/or operations organized through direct and/or operational output--input interconnections and/or output--parameter interconnections. Its transformations are parameterized transformations of the two kinds defined in \S3.2 and are defined within the value algebra, whereas the operational blocks required by its operational output--input interconnections are non-parameterized vector-to-vector functions that perform architecture-specific functional roles through mathematical functions}
\[
F:
\overline{\mathbb R}^{\,n}
\longrightarrow
\overline{\mathbb R}^{\,n}
\]
\emph{supplied by particular mathematical instantiations.}

This characterization remains deliberately \emph{semantic-free}. It specifies how values can be transformed and how transformations and operations can be coordinated, but it does not yet determine what vectors, parameter structures, or transformations mean.

\section{An Intensional--Extensional Formal Semantics for Representational Vectors}

The mathematical framework introduced in Section~3 specifies how values are organized and transformed, but it is deliberately neutral about their interpretation. We now add a semantic level by associating vector and matrix positions with properties and, in the case of representational vectors, by assigning the vector an extensional reference. The resulting semantics is intensional and extensional at the same time, but in different respects: properties are intensionally individuated, whereas representational vectors refer extensionally to objects. This distinction will also allow us to clarify why the semantic interpretation of parameter vectors and matrices differs from that of vectors representing objects.

\subsection{Objects, values, and properties}

Let $D$ be a fixed domain of objects, and let \(\mathcal P\) be a domain of \emph{intensional properties} that may be used to interpret vector and matrix positions.

Properties are not identified with their extensions, nor is an extension assigned to an intensional property once and for all. It is determined relative to a specific occurrence of that property in a position of a representational vector; the same property may therefore have different extensions across occurrences, and distinct properties may coincide extensionally in particular occurrences.

The property domain contains a distinguished \emph{null property} \(\nullprop.\) The null property interprets a position of a representational vector or a parameter structure if and only if the blank value $\blank$ occupies that position.

For the interpretation of parameter structures, we also introduce the distinguished \emph{preservation property} \(\wwstar\) and the extended parameter-interpretation domain
\[
\mathcal P_{\wwstar}
:=
\mathcal P
\cup
\{\wwstar\}.
\]
The preservation property is not an ordinary property of objects in $D$ and is therefore not assigned an occurrence-relative extension of the kind defined for properties occurring in representational vectors. It is a special interpretive element used only for parameter structures. Its transformational role will be defined in Section~5.

\subsection{Intensional interpretation of vector positions}

Consider a vector
\[
\boldsymbol v
=
(v_1,\ldots,v_m)
\in
\overline{\mathbb R}^{\,m}.
\]
An \emph{intensional interpretation} of its positions is an ordered $m$-tuple of properties
\[
\boldsymbol P
:=
(P_1,\ldots,P_m),
\qquad
P_i\in\mathcal P.
\]
Thus, each position $i$ is associated both with a value \(v_i\in\overline{\mathbb R}\) and with an intensional property \(P_i\in\mathcal P.\) These two assignments must be kept distinct. The first belongs to the value structure developed in Section~3; the second belongs to the intensional interpretation of the vector.

In particular, whenever \(P_i=\nullprop,\) the corresponding value is \(v_i=\blank.\) 
At this stage, however, the association of properties with vector positions does not yet make the vector a representation of an object. For that, an extensional component of the semantics is required.

\subsection{Extensional reference and features}

Every vector used as a representation has a fixed extensional interpretation. If $\boldsymbol v$ refers to the object $a\in D$, we write \(\operatorname{Ext}(\boldsymbol v) := a.\) 
Suppose that position $i$ of $\boldsymbol v$ is interpreted by $P_i$. The \emph{position- and vector-relative extension} of $P_i$ is then the function
\[
\operatorname{Ext}_{\boldsymbol v,i}(P_i):
\{a\}
\longrightarrow
\overline{\mathbb R}
\]
defined by
\[
\operatorname{Ext}_{\boldsymbol v,i}(P_i)(a)
:=
v_i.
\]

The relativization to both vector and position is essential. If the same property occurs at two positions of one vector, \(P_i=P_k=P\) with \(i\neq k\), while \(v_i\neq v_k\), then \(\operatorname{Ext}_{\boldsymbol v,i}(P)\) and \(\operatorname{Ext}_{\boldsymbol v,k}(P)\) are distinct functions because their values at $a$ differ. The same holds across different representational vectors referring to $a$. Thus identity of an intensional property does not require identity of its extensions across occurrences.

This also fixes the terminology used throughout the paper. \emph{Property} refers to the intensional entity associated with a vector or matrix position. \emph{Feature} is used when such a property is considered as a characteristic of a represented object, or when we refer to the value it takes in a particular representational occurrence. Thus, $P_i$ may be considered a feature of $a$, while $v_i$ is the corresponding feature value in position $i$ of $\boldsymbol v$.

A vector endowed both with an intensional interpretation \(\boldsymbol P = (P_1,\ldots,P_m)\) and with an extensional interpretation \(\operatorname{Ext}(\boldsymbol v)=a\) will be called a \emph{representational vector}. In the terminology inherited from TR1, it is a \emph{transformee concept}.

\subsection{The interpretation of parameter vectors and matrices}

Parameter vectors and matrices also admit an intensional interpretation of their positions, but their semantic structure differs in one essential respect from that of representational vectors.

Let
\[
\boldsymbol p
=
(p_1,\ldots,p_n)
\in
\overline{\mathbb R}_{\wstar}^{\,n}
\]
be a parameter vector. An intensional interpretation of its positions is an ordered tuple
\[
\boldsymbol Q
:=
(Q_1,\ldots,Q_n)
\in
\mathcal P_{\wwstar}^{\,n}.
\]
Thus, each parameter position associates a parameter value \(p_i\in \overline{\mathbb R}_{\wstar}\) with an interpretive element \(Q_i\in \mathcal P_{\wwstar}.\) 
Similarly, let
\[
\boldsymbol M
=
(m_{ij})
\in
\overline{\mathbb R}_{\wstar}^{\,m\times n}.
\]
An intensional interpretation of its positions assigns an element \(P_{ij} \in \mathcal P_{\wwstar}\) to each position $(i,j)$, producing an interpretive matrix \(\boldsymbol\Pi := (P_{ij}).\) 
These are \emph{position-wise associations}, not evaluation relations between parameter values and properties. In particular, the special parameter value $\wstar$ is not a value taken by an ordinary property on an object in $D$. Its presence therefore does not extend the codomain of ordinary property extensions beyond $\overline{\mathbb R}$.

The following compatibility conditions apply to every position of a parameter structure. The null property $\nullprop$ interprets that position if and only if the blank value $\blank$ occupies it. Thus, for a matrix position $(i,j)$,
\[
P_{ij}=\nullprop
\quad\Longleftrightarrow\quad
m_{ij}=\blank.
\]
If \(P_{ij}\in\mathcal P\setminus\{\nullprop\},\) then \(m_{ij}\in\mathbb R.\) If instead \(P_{ij}=\wwstar,\) then \(m_{ij}\in\mathbb R\cup\{\wstar\}.\) The same conditions apply, \emph{mutatis mutandis}, to the corresponding positions of parameter vectors.

What is absent in parameter structures is the extensional interpretation characteristic of representational vectors. A parameter vector or matrix does not, as such, refer to an object \(a\in D.\) Consequently, there is no vector extension analogous to \(\operatorname{Ext}(\boldsymbol v)=a,\) and no occurrence-relative property extension analogous to \(\operatorname{Ext}_{\boldsymbol v,i}(P_i).\) The values occupying parameter positions therefore do \emph{not} represent features of an object.

A parameter vector or matrix is a parameter structure only relative to the transformation it parameterizes: \(\boldsymbol M\) is structurally associated with \(L_{\boldsymbol M}^{(\star^{+},\diamond^{\times})}(\boldsymbol x)\), and \(\boldsymbol p\) with \(C_{\boldsymbol p}^{\diamond^{\times}}(\boldsymbol x)\). Section~5 gives this association semantic content by interpreting parameter positions relative to the properties of an arbitrary argument of the corresponding transformation.

\subsection{Formal semantics, representation, and grounding}

The preceding construction distinguishes three components in the semantics of a representational vector: \(i\longmapsto P_i,\) \[
\operatorname{Ext}(\boldsymbol v):=a,
\]
and
\[
\operatorname{Ext}_{\boldsymbol v,i}(P_i)(a):=v_i.
\]
The first supplies the intensional interpretation of a position, the second the extensional interpretation of the vector, and the third \emph{defines} the extension of the property relative to its occurrence in that position of that vector.

These formal assignments embody a structural conception of representation: a representational vector is treated as an object-referential structure whose positions are interpreted by properties with occurrence-relative values. Nevertheless, they are not themselves a philosophical theory of representation: \(i\mapsto P_i\) does not explain why $P_i$ is the appropriate interpretation, and \(\operatorname{Ext}(\boldsymbol v):=a\) does not explain what grounds the reference to $a$. The value $v_i$ is related to $a$ through \(\operatorname{Ext}_{\boldsymbol v,i}(P_i)(a):=v_i\); parameter values lack this relation to objects. Hence
\[
\boxed{
\text{formal semantic interpretation}
\neq
\text{representational relation}
\neq
\text{metasemantic grounding}.
}
\]
The present section specifies the first while distinguishing intensional property identity, extensional vector reference, and \emph{occurrence-relative property extension}; questions concerning the grounding of these assignments remain metasemantic. Representational vectors possess all three components, whereas parameter structures retain only an intensional interpretation of their positions.

The main result of the present section can therefore be summarized as follows.

\begin{quote}
\textbf{C1 --- Intensional--extensional formal semantics.}
\emph{A representational vector is semantically characterized by an extensional interpretation $\operatorname{Ext}(\boldsymbol v):=a$ together with an intensional interpretation $\boldsymbol P=(P_1,\ldots,P_n)$ of its positions. For each occurrence of $P_i$ in position $i$ of $\boldsymbol v$, its extension is defined relative to that vector and position by}
\[
\operatorname{Ext}_{\boldsymbol v,i}(P_i):
\{a\}\to\overline{\mathbb R},
\qquad
\operatorname{Ext}_{\boldsymbol v,i}(P_i)(a):=v_i.
\]
\emph{Parameter vectors and matrices may likewise receive intensional interpretations of their positions, but lack an extensional interpretation as representations of objects.}
\end{quote}

\section{Transformational Semantics and Transforming Concepts}

Section~4 introduced an intensional--extensional semantics for representational vectors and a non-object-referential interpretation of parameter structures. We now extend the semantics from vectors to parameterized transformations. The central step is to coordinate the value algebra of Section~3 with an algebra of properties. Parameter structures receive interpretations relative to the properties interpreting their arguments, and the resulting property-level maps are paired with the corresponding transformations of values. This provides the formal basis for the notion of a \emph{transforming concept}.

\subsection{The property algebra}

The special elements introduced in the preceding sections now acquire their transformational role.

At the value level, the preserving parameter value \(\wstar\) is a right identity for the operation occupying the multiplicative role: \(x\diamond^{\times}\wstar = x.\) At the property level, the preservation property \(\wwstar\) plays the corresponding property-preserving role.

We define
\[
\odot:
\mathcal P
\times
\mathcal P_{\wwstar}
\longrightarrow
\mathcal P,
\]
called \emph{projection}, by
\[
P\odot Q
:=
\begin{cases}
Q, & Q\in\mathcal P,\\[2mm]
P, & Q=\wwstar.
\end{cases}
\]
Hence, \(P\odot\wwstar = P.\) Thus, \(\wwstar\) is a right identity for projection.

We also assume a binary operation
\[
\oplus:
\mathcal P\times\mathcal P
\longrightarrow
\mathcal P,
\]
called \emph{property sum}, satisfying, for every \(P\in\mathcal P\),
\[
P\oplus P=P,
\qquad
\nullprop\oplus P=P,
\qquad
P\oplus\nullprop=P,
\]
and, for distinct non-null properties \(P_1\) and \(P_2\),
\[
P_1\oplus P_2\neq P_1
\qquad\text{and}\qquad
P_1\oplus P_2\neq P_2.
\]
Property sum is therefore idempotent, and the null property is a two-sided identity. The sum of two distinct non-null properties is a further property distinct from either of them.

No associativity assumption is imposed. Accordingly, finite property sums are defined by left association:
\[
\bigoplus_{i=1}^{1}Q_i:=Q_1,
\qquad
\bigoplus_{i=1}^{k+1}Q_i
:=
\left(
\bigoplus_{i=1}^{k}Q_i
\right)\oplus Q_{k+1}.
\]

The value and property algebras exhibit a parallel division of labor:
\[
\boxed{
\begin{array}{ccc}
\text{value level}
&&
\text{property level}
\\[1mm]
x_i\diamond^{\times}m_{ij}
&&
P_i\odot P_{ij}
\\[1mm]
\star^{+}
&&
\oplus .
\end{array}}
\]
The idempotence of \(\oplus\) is especially important: repeated contributions interpreted by the same property \(Q\) satisfy \(Q\oplus\cdots\oplus Q=Q\), although their numerical contributions remain distinct. This will be central to the analysis of attention in Section~8.

\subsection{Relative interpretations and second-level interpretations}

We can now make explicit the \emph{semantic content} of the structural association between parameter structures and argument vectors introduced in \S4.4.

Let
\[
\boldsymbol P
:=
(P_1,\ldots,P_m)
\in
\mathcal P^m
\]
be the interpretation of an arbitrary argument of the matrix-parameterized transformation
\[
L_{\boldsymbol M}^{(\star^{+},\diamond^{\times})},
\]
where
\[
\boldsymbol M
=
(m_{ij})
\in
\overline{\mathbb R}_{\wstar}^{\,m\times n}.
\]
An interpretation of \(\boldsymbol M\) \emph{relative to} \(\boldsymbol P\) is a property matrix
\[
\boldsymbol\Pi_{\boldsymbol P}
:=
(P_{ij})
\in
\mathcal P_{\wwstar}^{\,m\times n}.
\]
The subscript \(\boldsymbol P\) indicates that the interpretation of the parameter matrix is not semantically independent of the interpretation of an argument of the corresponding transformation. Each parameter position \((i,j)\) is interpreted relative to the property \(P_i\) associated with the corresponding position of that argument.

For every position \((i,j)\), the property-level contribution is \(P_i\odot P_{ij},\) in parallel with the value-level contribution \(x_i\diamond^{\times}m_{ij}.\) 
Each column \(j\) of \(\boldsymbol\Pi_{\boldsymbol P}\) defines an interpretive \emph{point of view}. Intuitively, point of view \(j\) specifies, for each argument property \(P_i\), whether a property is implicated, which property it is, and to what extent the implication holds.

More specifically, when \(P_{ij} \in \mathcal P\setminus\{\nullprop\},\) the implicated property is \(P_{ij}\) itself, since \(P_i\odot P_{ij} = P_{ij}.\) The parameter value \(m_{ij}\) specifies to what extent the implication holds. If, in particular, \(P_{ij}=P_i,\) the implication is identical.

When \(P_{ij}=\wwstar,\) the implicated property is instead \(P_i\), so that the implication is necessarily identical, again to the extent specified by \(m_{ij}\), since \(P_i\odot\wwstar = P_i.\) 
Finally, when \(P_{ij}=\nullprop,\) no property is implicated by \(P_i\) according to point of view \(j\), as \(P_i\odot\nullprop = \nullprop.\) That point of view makes no inferential contribution from \(P_i\). By the compatibility condition of \S4.4, this occurs exactly when \(m_{ij}=\blank,\) and therefore \(x_i\diamond^{\times}\blank = \blank.\) Thus, the absence of an implication at the property level corresponds to the absence of a numerical contribution at the value level.

When an implication is present, the parameter value \(m_{ij}\) associated with the implicated property does not by itself contribute to the value of the resulting property. Intuitively, \(m_{ij}\) specifies the strength of the implication from \(P_i\) to the implicated property, according to point of view \(j\). For a representational argument, the effective contribution of this implication must therefore also depend on the value \(x_i\) with which the implicating property \(P_i\) holds of the represented object. For this reason, the contribution to the resulting value is not \(m_{ij}\) alone, but \(x_i\diamond^{\times}m_{ij}.\) The operation occupying the multiplicative role thus combines the strength with which the implicating property holds with the strength of the implication itself.

At the property level, the corresponding contribution is the implicated property \(P_i\odot P_{ij},\) whereas at the value level its contribution is \(x_i\diamond^{\times}m_{ij}.\) 
The resulting property associated with output position \(j\) is obtained by combining the properties implicated according to point of view \(j\):
\[
R_j
:=
\bigoplus_{i=1}^{m}
\left(
P_i\odot P_{ij}
\right).
\]
Writing \(\boldsymbol R := (R_1,\ldots,R_n),\) we define the corresponding \emph{property-level map} by
\[
L_{\boldsymbol\Pi_{\boldsymbol P}}^{(\oplus,\odot)}
(\boldsymbol P)
:=
\boldsymbol R.
\]

At the value level,
\[
\boldsymbol y
:=
L_{\boldsymbol M}^{(\star^{+},\diamond^{\times})}
(\boldsymbol x),
\]
where
\[
y_j
:=
\starsum_{i=1}^{m}
\left(
x_i\diamond^{\times}m_{ij}
\right).
\]
The value-level and property-level constructions therefore proceed in parallel:
\[
\boxed{
\begin{aligned}
y_j
&:=
\starsum_{i=1}^{m}
\left(
x_i\diamond^{\times}m_{ij}
\right),
\\[2mm]
R_j
&:=
\bigoplus_{i=1}^{m}
\left(
P_i\odot P_{ij}
\right).
\end{aligned}}
\]
The resulting property \(R_j\) is thus obtained by combining the properties implicated according to point of view \(j\), while the corresponding resulting value \(y_j\) is obtained by combining their numerical contributions through \(\star^{+}\).

When \(\boldsymbol x\) is a representational argument referring to \(a\in D\), we further assume that the resulting vector \(\boldsymbol y\) is itself a representational vector referring to the same object \(a\):
\[
\operatorname{Ext}(\boldsymbol x)
=
\operatorname{Ext}(\boldsymbol y)
=
a.
\]
If \(\boldsymbol R\) interprets \(\boldsymbol y\), then by the definition introduced in \S4.3,
\[
\operatorname{Ext}_{\boldsymbol y,j}(R_j)(a)
=
y_j,
\qquad
j=1,\ldots,n.
\]
The value-level transformation and the property-level map remain formally distinct. Their results are coordinated semantically insofar as \(\boldsymbol R\) provides the intensional interpretation of the resulting representational vector \(\boldsymbol y\), whose occurrence-relative property extensions are thereby fixed by its values.

We call \(\boldsymbol\Pi_{\boldsymbol P}\) a \emph{second-level interpretation} because projection admits a natural semantic interpretation in terms of implication. From point of view \(j\), \(P_i\odot P_{ij}\) can be read as the property implicated by \(P_i\): when \(P_{ij}\in\mathcal P\setminus\{\nullprop\}\), right-projection yields \(P_{ij}\); when \(P_{ij}=P_i\) or \(P_{ij}=\wwstar\), the implication is identical; and the null case expresses the absence of an implication. On this interpretation, when \(P_{ij}\neq P_i\), the implicated property may naturally be regarded as more abstract than \(P_i\), since it expresses what \(P_i\) implies according to point of view \(j\) rather than merely reproducing \(P_i\). The subsequent property sum may then compose several such implicated properties into \(R_j\).

By contrast, property sum admits a natural interpretation as \emph{abstraction-level preserving}: the property resulting from the combination can be regarded as neither less abstract than any of its component properties nor more abstract than each of them, since their combination introduces no further implication and eliminates none of those already expressed by its components.

On this reading, the framework provides a formal-semantic explanation of the familiar capacity of neural networks to produce progressively more abstract representations: abstraction arises through the replacement, by projection, of argument properties with distinct properties implicated from different points of view, while property sum composes those implicated properties without itself producing or destroying abstraction.

The same semantic construction applies to vector-parameterized componentwise transformations, but in a structurally restricted form. Let
\[
\boldsymbol v
=
(v_1,\ldots,v_n)
\in
\overline{\mathbb R}_{\wstar}^{\,n}
\]
parameterize
\[
C_{\boldsymbol v}^{\diamond^{\times}}
:
\overline{\mathbb R}^{\,n}
\longrightarrow
\overline{\mathbb R}^{\,n},
\]
and let
\[
\boldsymbol P
=
(P_1,\ldots,P_n)
\in
\mathcal P^n
\]
interpret an arbitrary argument. An interpretation of \(\boldsymbol v\) relative to \(\boldsymbol P\) is a tuple
\[
\boldsymbol Q_{\boldsymbol P}
:=
(Q_1,\ldots,Q_n)
\in
\mathcal P_{\wwstar}^{\,n}.
\]

Each position \(j\) of the parameter vector defines a point of view restricted to the corresponding argument property \(P_j\), rather than ranging over all \(P_i\) as a matrix column can. Thus \(Q_j\in\mathcal P\setminus\{\nullprop\}\) implicates \(Q_j\), with identical implication when \(Q_j=P_j\); \(Q_j=\wwstar\) preserves \(P_j\); and \(Q_j=\nullprop\) implicates no property. The parameter value \(v_j\) specifies the extent of the implication through the value contribution \(x_j\diamond^{\times}v_j\). In particular, \(Q_j=\wwstar\) together with \(v_j=\wstar\) preserves both property and value.
The property-level map is consequently
\[
C_{\boldsymbol Q_{\boldsymbol P}}^{\odot}
(\boldsymbol P)
:=
\left(
P_1\odot Q_1,\ldots,
P_n\odot Q_n
\right),
\]
in parallel with the value-level transformation
\[
C_{\boldsymbol v}^{\diamond^{\times}}
(\boldsymbol x)
:=
\left(
x_1\diamond^{\times}v_1,\ldots,
x_n\diamond^{\times}v_n
\right).
\]
Thus, for each point of view \(j\),
\[
\boxed{
\begin{array}{ccc}
P_j &\longrightarrow& P_j\odot Q_j,\\[1mm]
x_j &\longrightarrow& x_j\diamond^{\times}v_j.
\end{array}}
\]

A vector-parameterized transformation therefore differs from a general matrix-parameterized transformation in one important semantic respect. A matrix-defined point of view may draw implications from several or all argument properties and combine the resulting implicated properties into a single output property. A vector-defined point of view is restricted to the corresponding argument property \(P_j\). It may preserve that property, replace it by a distinct property implicated by it, or imply no property at all, but it cannot combine implications originating from different argument positions.

The diagonal representation makes this restriction explicit. Let
\[
\boldsymbol D(\boldsymbol Q_{\boldsymbol P})
:=
\begin{pmatrix}
Q_1 & \nullprop & \cdots & \nullprop\\
\nullprop & Q_2 & \cdots & \nullprop\\
\vdots & \vdots & \ddots & \vdots\\
\nullprop & \nullprop & \cdots & Q_n
\end{pmatrix}.
\]
In column \(j\), all off-diagonal positions are null, so point of view \(j\) is semantically restricted to the corresponding argument property \(P_j\). Consequently,
\[
L_{\boldsymbol D(\boldsymbol Q_{\boldsymbol P})}^{(\oplus,\odot)}
(\boldsymbol P)
=
C_{\boldsymbol Q_{\boldsymbol P}}^{\odot}
(\boldsymbol P).
\]
The diagonal correspondence established at the value level in \S3.2 therefore has a precise semantic counterpart: \emph{vector parameterization corresponds to a family of interpretive points of view, each restricted to the corresponding argument property.}

\subsection{Transforming concepts}

We can now state formally the notion introduced provisionally in \S2.

For a fixed argument interpretation \(\boldsymbol P \in \mathcal P^m\) and a corresponding relative interpretation \(\boldsymbol\Pi_{\boldsymbol P}\) of the parameter matrix \(\boldsymbol M\), a \emph{matrix-parameterized transforming concept} is identified with the ordered pair
\[
\boxed{
\left(
L_{\boldsymbol M}^{(\star^{+},\diamond^{\times})},
\;
L_{\boldsymbol\Pi_{\boldsymbol P}}^{(\oplus,\odot)}
\right).
}
\]
The first member is the value-level transformation, \(\boldsymbol x \longmapsto \boldsymbol y,\) while the second is the corresponding property-level map, \(\boldsymbol P \longmapsto \boldsymbol R.\) Together they determine
\[
(\boldsymbol x,\boldsymbol P)
\longmapsto
(\boldsymbol y,\boldsymbol R).
\]

For vector-parameterized componentwise transformations, the corresponding transforming concept is
\[
\boxed{
\left(
C_{\boldsymbol v}^{\diamond^{\times}},
\;
C_{\boldsymbol Q_{\boldsymbol P}}^{\odot}
\right).
}
\]

A transforming concept therefore requires more than mathematical parameterization. It requires a relative interpretation of the parameter structure and a corresponding property-level map. This makes precise the distinction between \emph{mathematical parameterization} and \emph{conceptual parameterization}.

Operational blocks are not transforming concepts: they have no parameter structure with a relative, second-level interpretation and hence no parameter-induced property-level map. Since they nevertheless map representational vectors to vectors that may become later arguments, we require them to preserve the input's extensional reference and ordered intensional interpretation. Thus, if \(\boldsymbol y=F(\boldsymbol x)\), both vectors refer to the same object $a$ and are interpreted by the same tuple \(\boldsymbol P\), although \(\operatorname{Ext}_{\boldsymbol x,i}(P_i)(a)=x_i\) and \(\operatorname{Ext}_{\boldsymbol y,i}(P_i)(a)=y_i\) may differ.

Preserving the complete ordered tuple also preserves its arity, so operational blocks are dimension-preserving functions
\[
\boxed{F:\overline{\mathbb R}^{\,n}\longrightarrow\overline{\mathbb R}^{\,n}.}
\]
They may modify values and occurrence-relative extensions, but not the intensional interpretation. \emph{An operational block is therefore intensionally invariant.}

The central distinction can therefore be summarized as follows:
\[
\boxed{
\begin{array}{c}
\text{parameterized transformation}\\
+\ \text{relative second-level interpretation}\\
\Downarrow\\
\text{transforming concept},
\end{array}
\qquad
\begin{array}{c}
\text{non-parameterized operational block}\\
+\ \text{intensional invariance}\\
\Downarrow\\
\text{not a transforming concept}.
\end{array}}
\]

\subsection{Intensional invariance and dimensionality}

The same notion of intensional invariance can be extended to transforming concepts. In this case, however, intensional invariance is not required by their semantic structure: a transforming concept may either preserve or change the ordered interpretation of its argument.

Let \(\boldsymbol P = (P_1,\ldots,P_m)\) be the interpretation of the argument and \(\boldsymbol R = (R_1,\ldots,R_n)\) the interpretation of the resulting vector. A transforming concept is \emph{intensionally invariant relative to \(\boldsymbol P\)} if and only if
\[
\boxed{
\boldsymbol R
=
\boldsymbol P.
}
\]
Otherwise, it is \emph{intensionally changing}.

Since equality of ordered tuples requires equality of their arities,
\[
\boldsymbol R=\boldsymbol P
\quad\Longrightarrow\quad
m=n.
\]
Consequently,
\[
\boxed{
m\neq n
\quad\Longrightarrow\quad
\text{intensional invariance is impossible}.
}
\]
When \(m=n\), dimensional preservation alone does not guarantee intensional invariance: the transformation may still produce an interpretation \(\boldsymbol R\neq\boldsymbol P.\) 
Let now
\[
\boldsymbol Q_{\boldsymbol P}
=
(\wwstar,\ldots,\wwstar).
\]
The componentwise transforming concept
\[
\left(
C_{\boldsymbol v}^{\diamond^{\times}},
\;
C_{\boldsymbol Q_{\boldsymbol P}}^{\odot}
\right)
\]
is then intensionally invariant because \(P_i\odot\wwstar = P_i\) for every \(i\), and therefore
\[
C_{\boldsymbol Q_{\boldsymbol P}}^{\odot}
(\boldsymbol P)
=
\boldsymbol P.
\]

The corresponding parameter values need not preserve the numerical values. In general,
\[
C_{\boldsymbol v}^{\diamond^{\times}}(\boldsymbol x)
\neq
\boldsymbol x
\]
may hold while
\[
C_{\boldsymbol Q_{\boldsymbol P}}^{\odot}
(\boldsymbol P)
=
\boldsymbol P.
\]
Hence, \emph{intensional invariance does not imply value preservation}.

The preserving value \(\wstar\) permits both forms of preservation to coincide: if \(\boldsymbol v=(\wstar,\ldots,\wstar)\) and \(\boldsymbol Q_{\boldsymbol P}=(\wwstar,\ldots,\wwstar)\), then \(C_{\boldsymbol v}^{\diamond^{\times}}(\boldsymbol x)=\boldsymbol x\) and \(C_{\boldsymbol Q_{\boldsymbol P}}^{\odot}(\boldsymbol P)=\boldsymbol P\), so \((\boldsymbol x,\boldsymbol P)\mapsto(\boldsymbol x,\boldsymbol P)\).

The structural possibilities can thus be summarized as follows:
\[
\boxed{
\begin{array}{lll}
m=n,\ \boldsymbol R=\boldsymbol P
&\Rightarrow
&\text{dimension-preserving and intensionally invariant},
\\[1mm]
m=n,\ \boldsymbol R\neq\boldsymbol P
&\Rightarrow
&\text{dimension-preserving and intensionally changing},
\\[1mm]
m\neq n
&\Rightarrow
&\text{dimension-changing and necessarily intensionally changing}.
\end{array}}
\]
The second case is especially important because it shows that a change of intensional interpretation need not be accompanied by a change of dimensionality. Residual additions will provide an important Transformer case in Section~8.

\subsection{The genesis of values and interpretations}

The preceding analysis answers the \emph{constitutive} question of what makes a parameterized transformation a transforming concept. We distinguish from it a \emph{genetic} question concerning how parameter values and interpretations arise, addressed here only schematically, and a \emph{metasemantic} question concerning what grounds the relevant interpretive and representational relations. The latter connects the framework with causal/informational, functional/inferential, environmental, and other accounts of content determination, among which we do not adjudicate.

For the limited genetic classification developed below, we distinguish
\[
G_{\mathrm{val}}
:=
\text{genesis of the parameter values},
\]
from
\[
G_{\mathrm{int}}
:=
\text{genesis of their intensional interpretation}.
\]

Let \(E := \text{environmental interaction},\) and
\[
T
:=
\text{theoretical/architectural determination}.
\]

The genetic dimensions represented by \(G_{\mathrm{val}}\) and \(G_{\mathrm{int}}\) are distinct from the distinction between intensional invariance and intensional change established in \S5.4. Whether a transforming concept is intensionally invariant depends on the relation between its argument and output interpretations, not on how either its parameter values or its parameter interpretation originated.

The following table is constructed from distinctions internal to the theory developed above. Its fourth column is added only to indicate Transformer instances of the corresponding configurations. The table is not intended as an exhaustive combinatorial taxonomy of every logically possible assignment of \(E\) and \(T\).

\begin{center}
\small
\renewcommand{\arraystretch}{1.3}
\begin{tabular}{>{\centering\arraybackslash}p{0.12\textwidth}
                >{\centering\arraybackslash}p{0.12\textwidth}
                >{\raggedright\arraybackslash}p{0.22\textwidth}
                >{\raggedright\arraybackslash}p{0.42\textwidth}}
\toprule
$G_{\mathrm{val}}$ & $G_{\mathrm{int}}$ & \textbf{Intensional effect} & \textbf{Illustrative Transformer cases}\\
\midrule
$E$ & $E$ & Intensionally changing & Learned matrices such as $W^Q,W^K,W^V$; dynamically generated parameter structures such as $K^\top$ and $V$\\
$E$ & $T$ & Intensionally invariant & Learned bias and scaling vectors; trainable positional vectors\\
$T$ & $T$ & Intensionally invariant & Sinusoidal positional vectors; masking, rescaling, and temperature\\
$T$ & $T$ & Intensionally changing & ---\\
\bottomrule
\end{tabular}
\end{center}

The first three configurations have Transformer instances; the fourth is formally permitted but has no clear one. The second and third rows are intensionally invariant for theoretical reasons: bias and scaling vectors, positional vectors, masking, rescaling, and temperature modify values without introducing a new interpretation, so their second-level interpretation contains \(\wwstar\) throughout. Their difference lies in value genesis---training in the second row, theoretical or architectural determination in the third.

The first row includes two modes of environmental dependence: learned matrices such as \(W^Q,W^K,W^V\) depend on training, whereas dynamically generated parameter structures acquire both values and second-level interpretations during inference from the current input. Thus \(E\) covers both modes without identifying them. The fourth row also makes explicit that theoretically or architecturally determined values and interpretations may nevertheless yield an intensionally changing transforming concept. Overall, the table adds a genetic classification without altering the constitutive semantics: genesis does not determine transforming-concept status or intensional invariance. Likewise, the static/dynamic distinction concerns genesis and temporal availability rather than constitutive semantic status.

The most significant results of this section can therefore be summarized as follows.

\begin{quote}
\textbf{C2 --- Transformational formal semantics.}
\emph{A parameterized transformation constitutes a transforming concept when its value-level transformation is coordinated with a property-level map determined by a relative, second-level interpretation of its parameter structure. The resulting transforming concept may be intensionally invariant or intensionally changing. Its constitutive semantic status is independent both of the genesis of its parameter values and interpretations and of which account, if any, is adopted to ground the relevant interpretive and representational relations. At the same time, by specifying precisely what is interpreted, at which level, how the different interpretive relations are connected, and how they contribute to the transformation of representational vectors, the present formal semantics provides the conceptual articulation required for posing metasemantic questions in a form adequate to the complexity of the systems under discussion.}
\end{quote}

\section{Semantic Composition in Coordinated Systems}

We now combine the architectural organization of Section~3 with the semantics of Sections~4--5. Rather than introducing a recursive semantics for arbitrary compositions, we identify the semantic conditions associated with the principal forms of coordination already distinguished architecturally.

\subsection{Interpretive uniformity in operations}

In \S3.3, an \emph{operation} was defined mathematically as the parallel organization of a finite family of transformations, where \emph{transformation} is used in the technical sense fixed in \S3.2 for the two kinds of parameterized transformations introduced there. At the semantic level developed in \S5, we can now consider the corresponding family of \emph{transforming concepts}.

Let
\[
T_k:
\overline{\mathbb R}^{\,m}
\longrightarrow
\overline{\mathbb R}^{\,n},
\qquad
k=1,\ldots,s,
\]
be the value-level transformations of such a family, and let
\[
\boldsymbol x_1,\ldots,\boldsymbol x_s
\in
\overline{\mathbb R}^{\,m}
\]
be their argument vectors. The $T_k$ need not be mathematically identical, and their parameter structures and relative second-level interpretations may also differ.

We call a parallel family of transforming concepts \emph{interpretively uniform} when all its argument vectors share the same ordered intensional interpretation \(\boldsymbol P = (P_1,\ldots,P_m),\) and all its resulting vectors share the same ordered intensional interpretation \(\boldsymbol R = (R_1,\ldots,R_n).\) 
If \(\boldsymbol y_k = T_k(\boldsymbol x_k),\) interpretive uniformity requires that each $\boldsymbol x_k$ be interpreted by $\boldsymbol P$ and each $\boldsymbol y_k$ by $\boldsymbol R$.

Corresponding positions can then be described by the uniform semantic pattern \(\boldsymbol P\mapsto\boldsymbol R\), even when the individual transforming concepts differ in value-level transformations, parameter values, or relative second-level interpretations. Thus common arities suffice for a mathematical operation, whereas a semantically coordinated operation additionally requires a common ordered interpretation of corresponding input and output positions. Positional-vector addition and masking provide elementary Transformer examples.

\subsection{Direct interconnections and semantic transmission}

Direct interconnections introduce no further mathematical function. Their semantic effect therefore depends only on their architectural form. In a \emph{standard direct interconnection}, an output vector becomes a later argument with its values, extensional reference, and ordered interpretation unchanged. A \emph{routing interconnection} may regroup or reorder output vectors, in which case the interpretive organization follows the same pattern. Transformer Concat illustrates this: output vectors from different heads for the same token are concatenated together with their property tuples and retain that token referent. Routing alone, however, cannot determine the reference of a vector formed from representations with different referents; that would require additional semantic or architectural information. A direct interconnection therefore transmits already interpreted vectors or reorganizes them according to its architectural form.

\subsection{Operational interconnections and intensional invariance}

Operational interconnections differ because an operational block modifies transmitted values. By \S5.3 it is nevertheless intensionally invariant: in \(T_1\to F\to T_2\), if the output \(\boldsymbol x\) of $T_1$ refers to $a$ and is interpreted by \(\boldsymbol P\), then \(F(\boldsymbol x)\) retains both while its values and occurrence-relative extensions may change. Thus no new interpretive level is introduced between source and destination; unlike a parameterized transformation, the value modification is not accompanied by a change of ordered intensional interpretation.

\subsection{Output--parameter interconnections and second-level interpretation}

Output--parameter interconnections introduce a different form of semantic composition. Here an output of a preceding transformation does not become the argument of a subsequent transformation, but all or part of its parameter structure.

Schematically, \(T_1(\boldsymbol x) = \boldsymbol p,\) followed by \(T_{\boldsymbol p}(\boldsymbol y).\) The same numerical structure therefore participates successively in two distinct semantic roles. As an output of $T_1$, $\boldsymbol p$ may be a representational vector with an extensional reference and an intensional interpretation of its positions. When it is subsequently used as a parameter structure, however, it is no longer interpreted in that role. Qua parameter structure, it receives instead a relative, second-level interpretation with respect to the intensional interpretation of the argument of the destination transformation.

Thus, the transition from output to parameter is not merely a numerical reassignment. It is also a \emph{change of semantic role}. The output--parameter interconnection connects a first-level representational stage with the second-level interpretive structure required for conceptual parameterization.

An output--parameter interconnection changes the semantic role of the transmitted numerical structure without necessarily erasing its \emph{referential provenance}. Qua parameter structure, the resulting vector or matrix has no extensional reference of its own. However, proper parts of the parameter structure may retain their relation of provenance to the representational vectors from which they were generated, or even the entire structure as a whole may retain such a relation. This provenance may remain semantically or architecturally relevant in the new role of the parameter structure. It may contribute to the second-level interpretation acquired by the parameter structure, or to the coordination between dynamically generated parameters and the argument vectors to which they are subsequently applied. The precise role of such referential provenance depends on the architecture of the relevant output--parameter interconnection.

With this caveat, the semantic coordination required by an output--parameter interconnection should not be understood as the preservation of the output's previous representational interpretation. It consists instead in providing the parameter structure, in its new role, with the relative, second-level interpretation required by the transformation in which it now functions as parameter.

This distinction will become especially important in the analysis of attention and residual connections in \S8.

\subsection{Semantic coordination at system level}

At the system level, semantic coordination results from the compatibility of the local semantic structures established in the preceding sections with the architectural relations defined in \S3. Representational vectors satisfy the intensional--extensional semantics summarized in C1, while transformations receive the transformational semantics summarized in C2. Operations additionally require interpretive uniformity.

Direct interconnections preserve semantic coordination in ways determined by their architectural form, as illustrated paradigmatically by standard token-wise interconnections and routing structures such as Concat in the Transformer. Operational interconnections preserve the intensional interpretation through intensionally invariant operational blocks, while output--parameter interconnections satisfy the semantic conditions established in \S6.4.

A coordinated system is therefore \emph{semantically coordinated} when these local semantic constraints are mutually compatible with its architecture of transformations, operations, operational blocks, and interconnections. No further global semantic mechanism is required: system-level coordination results from the way in which locally interpreted vectors, transformations, and parameter structures are connected according to the functional organization of the system.

At this point, the framework specifies both the mathematical and the semantic coordination of the systems under consideration. The next question concerns a different level of description: which aspects of this organization belong to the \emph{abstract functional architecture} of a system, and which belong instead to the particular mathematical instantiations of that architecture.

\section{Abstract Functional Architecture, Mathematical Instantiation, and Realization}

The preceding sections have characterized \emph{semantically coordinated systems of transformations} at a level more abstract than any of their particular mathematical instantiations. Their common algebraic-semantic structure has been specified independently of the particular numerical operations and mathematical functions through which it may be instantiated, while their coordination has been described in terms of transformations, operations, operational blocks, and interconnections. We can now make explicit this level of abstraction and distinguish it from the mathematical, mechanical, and physical levels associated with it.

\subsection{Abstract functional architecture}

An \emph{abstract functional architecture}, equivalently an \emph{abstract system type}, combines an \emph{architecture-independent algebraic-semantic core} with an \emph{architecture-specific functional organization}. The core comprises the value algebra of \S3 and the semantics of \S\S4--6, fixing the roles of contribution, combination, projection, property sum, and the semantic organization of vectors, parameters, transforming concepts, and coordination. The functional organization specifies the arrangement and arities of transformations and operations, the pattern of direct, operational, and output--parameter interconnections, and the sources, destinations, and functional roles of operational blocks.

Different abstract architectures therefore share the same core and differ in functional organization. The present paper specifies the core in detail while characterizing only some constraints on admissible organizations, thereby providing a common basis for a plurality of abstract functional architectures rather than a unique one.

\subsection{Mathematical instantiations}

An abstract functional architecture does not determine a unique mathematical instantiation. The algebraic-semantic core itself, however, exhibits an important asymmetry between its value and property components.

The domain of the \emph{value algebra} is fixed. Argument and resulting vectors range over \(\overline{\mathbb R},\) while parameter structures range over \(\overline{\mathbb R}_{\wstar}.\) What varies across mathematical instantiations are the operations occupying the abstract roles represented by
\[
\star^{+}
\qquad\text{and}\qquad
\diamond^{\times}.
\]
The constraints imposed on their underlying binary operations \(\star\) and \(\diamond\) are deliberately weak: apart from the specified behavior of their extensions with respect to the blank and preserving value, they are compatible with arbitrary binary operations on the real numbers. Different models of the value algebra may therefore assign mathematically very different operations to the same abstract roles. Since the value domain is fixed, variation between such models lies in the operations rather than in the domain on which they act.

The \emph{property algebra} has a different structure. Its domain \(\mathcal P\) is not fixed once and for all: different interpretations may employ different domains of intensional properties. The operations on those properties are constrained by the conditions introduced in \S5. Once \(\mathcal P\) is fixed, projection \(\odot\) is thereby fixed, whereas property sum \(\oplus\) remains only partially determined: for distinct non-null properties, its value must be a property distinct from either of them, but different operations may satisfy this condition. Different admissible property algebras may therefore differ both in their property domains and, for a fixed domain, in the way property sum is defined on distinct non-null properties.

The algebraic-semantic core consequently combines two complementary forms of model variation:
\[
\boxed{
\begin{array}{c|c|c}
& \text{domain} & \text{operations}\\
\hline
\text{value algebra} & \text{fixed} & \text{variable}\\
\text{property algebra} & \text{variable} & \odot\ \text{fixed},\ \oplus\ \text{constrained}.
\end{array}}
\]

The object domain \(D\) may of course vary between applications of the semantics. Such variation does not alter either of the two algebras or the semantic structure defined over them; it changes only the domain to which that semantics is applied.

At the architecture-specific level, further mathematical variation arises because a mathematical instantiation must supply particular functions
\[
F:
\overline{\mathbb R}^{\,n}
\longrightarrow
\overline{\mathbb R}^{\,n}
\]
for operational blocks whose abstract specification determines their functional roles, sources, destinations, and arities but not the mathematical functions performing those roles. Different functions may therefore instantiate the same architecture-specific functional role.

A mathematical instantiation of an abstract functional architecture is consequently specified by a model of its value algebra together with mathematical functions instantiating its architecture-specific functional roles. It does not determine a particular model of the property algebra or of the extensional--intensional semantics. Rather, it determines a class of admissible algebraic-semantic models sharing the same value-algebra model and the same formal semantic constraints, while potentially differing in their object domain, property domain, admissible property-sum operation, and extensional and intensional assignments.

\subsection{Functional-architectural invariance}

Two mathematical instantiations of the same abstract architecture may differ in numerical operations or mathematical functions while preserving the same algebraic-semantic structure and architecture-specific organization---the kinds and arities of transformations and operations, interconnection pattern, and operational roles, sources, and destinations. The system is \emph{functionally-architecturally invariant} across such instantiations.

This is not mathematical identity: mathematically different systems may instantiate the same abstract architecture, while the same function may belong to different abstract types when it occupies different roles or organizations. Abstract system type is therefore individuated by functional organization within the common algebraic-semantic framework, not by particular numerical operations or functions.

\subsection{Abstract system types and the Transformer}

The Transformer provides a paradigmatic illustration.

At the level considered here, the Transformer can be characterized as an abstract type of semantically coordinated system of transformations. It shares the common algebraic-semantic core developed above, while its architecture-specific functional organization determines its characteristic forms of parallel organization, direct and operational output--input interconnections, output--parameter interconnections, and the functional roles of its operational blocks.

The abstract Transformer type may admit different mathematical instantiations involving different mathematical choices. Conversely, other abstract system types share the same algebraic-semantic core while differing from the Transformer in their organization of transformations and operations, their patterns of interconnection, or the functional roles assigned to operational blocks.

The Transformer in its standard mathematical instantiation is used throughout this paper as a particularly important exemplar of one such type. Its role is illustrative rather than restrictive.

\subsection{Mathematical instantiation, mechanical realization, and the physical level}

Mathematical instantiation and \emph{mechanical realization} are distinct relations to an abstract type. A mathematical instantiation of the abstract functional system assigns particular mathematical operations and functions to its abstract roles. A mechanical realization instead specifies an \emph{idealized mechanical system} whose components and interactions realize the same algebraic-semantic core and the same architecture-specific functional organization.

Neither relation derives from the other. An abstract system type may admit different mathematical instantiations and different mechanical realizations independently. A particular mathematical instantiation of the abstract system does not determine which kind of mechanism realizes it, and a mechanical realization does not determine which mathematical instantiation is assigned to it.

Because every mechanical realization preserves the functional organization of the abstract system type it realizes, any mathematical instantiation of an abstract type \(A\) can also be assigned to any mechanical realization \(R\) of \(A\). Thus, if \(M\) is a mathematical instantiation of \(A\) and \(R\) mechanically realizes \(A\), the same \(M\) can be assigned to both \(A\) and \(R\):
\[
A\xrightarrow{M}A_M
\]
and, independently,
\[
A\longrightarrow R\xrightarrow{M}R_M.
\]
This branching permits two routes to the physical level.

A mathematically instantiated functional system \(A_M\) can be \emph{computationally implemented through computational simulation} without a mechanical realization. Here, \emph{computational simulation} is understood in the procedural sense of a method for implementing \(A_M\) on physical computers (cf. Humphreys 1990). We do not use \emph{simulation} here in the representational sense in which a simulation model stands in a distinct representational relation to a target. Thus,
\[
\boxed{\text{computational simulation}=\text{programming}+\text{numerical computation}}
\]
In this characterization, programming reproduces the functional architecture and numerical computation evaluates the mathematical operations and functions specified by \(M\). Thus
\[
A\xrightarrow{M}A_M
\xrightarrow{\substack{\text{programming}\\+\;\text{numerical computation}}}
\text{computational implementation}.
\]
The physical infrastructure need only execute this program and computation; it need not itself possess the mechanical organization of \(A\).

A mechanical realization provides the other route. Once the mathematical instantiation \(M\) is assigned to it, \(R_M\) specifies idealized components and interactions that can guide \emph{physical construction}:
\[
A\longrightarrow R\xrightarrow{M}R_M
\longrightarrow
\text{physical construction}.
\]
Thus a mathematically instantiated functional system can be physically implemented by simulation, whereas a mathematically instantiated mechanical realization can be materially constructed so that its components realize the mathematically instantiated mechanical organization.

For the Transformer, assigning the standard mathematical instantiation \(M_{\mathrm{std}}\) to the abstract type \(A_T\) yields a mathematically instantiated system that can be computationally implemented by programming its architecture and numerically evaluating the mathematical operations and functions specified by \(M_{\mathrm{std}}\). Independently, \(A_T\) admits a mechanical realization in interconnected groups of simple neural networks, which can be given the same \(M_{\mathrm{std}}\); other realizations, such as neuromorphic ones, may likewise be given the same mathematical instantiation while motivating different physical constructions. The structure is therefore not a linear hierarchy: mathematical instantiation and mechanical realization branch from the abstract type, leading respectively to computational implementation or, through a mathematically instantiated mechanical realization, to physical construction.

The present paper has concentrated on the \emph{abstract system level}. Its main contribution has been to identify the algebraic-semantic core shared by the class of systems considered here and to show how different architecture-specific functional organizations can be specified upon that core. Particular types of architecture, especially the Transformer architecture in its standard mathematical instantiation, have served to illustrate this level. A complete theory of mathematical instantiation, mechanical realization, computational implementation, or physical construction lies beyond the present scope.

The most significant results of this section can therefore be summarized as follows.

\begin{quote}
\textbf{C3 --- Functional-architectural invariance and routes to the physical level.}

\emph{Different abstract functional architectures share the same algebraic-semantic core and differ in their architecture-specific functional organization. Each abstract functional architecture may admit different mathematical instantiations and different mechanical realizations. Variations in particular numerical operations or mathematical functions do not constitute a change of abstract system type insofar as the common algebraic-semantic structure and the relevant organization of transformations, operations, interconnections, and functional roles are preserved.}

\emph{This distinction also reveals two routes from the abstract functional level to physical systems. A mathematically instantiated abstract functional system can be computationally implemented through computational simulation: programming reproduces its functional organization, while numerical computation evaluates the mathematical operations and functions specified by its mathematical instantiation, without requiring an intervening mechanical realization. Alternatively, a mechanical realization of the same abstract functional architecture can be given the same mathematical instantiation and then physically constructed so that material components and interactions realize its mathematically instantiated mechanical organization. Computational implementation and physical construction thus constitute distinct routes from abstract functional systems to the physical level.}
\end{quote}

\section{Attention and Residual Connections}

The preceding framework can now be used to analyze two characteristic processing patterns exemplified by the Transformer: attention heads and residual connections. The purpose is not to reproduce their standard mathematical description, but to identify the semantic organization determined by the local functional architectures of these modular patterns. The resulting explanations depend neither on the particular numerical operations through which they are mathematically instantiated in the standard Transformer nor on the overall functional architecture of the Transformer itself. They depend only on the abstract functional organization of the attention-head and residual-connection structures under consideration, together with the algebraic-semantic core shared by all semantically coordinated systems of transformations. They therefore apply to any abstract type of semantically coordinated system of transformations whose architecture includes structures of the same functional kinds.

\subsection{Attention scores and token-indexed points of view}

Consider an attention head operating on \(n\) representational vectors. Let
\[
Q,K\in\overline{\mathbb R}^{\,n\times d_k},
\qquad
V\in\overline{\mathbb R}^{\,n\times d_v},
\]
with rows
\[
\boldsymbol q_i,\boldsymbol k_i\in\overline{\mathbb R}^{\,d_k},
\qquad
\boldsymbol v_i\in\overline{\mathbb R}^{\,d_v},
\qquad
i=1,\ldots,n.
\]
The rows indexed by \(i\) originate from the corresponding token \(a_i\). By interpretive uniformity, the rows of each matrix share a common ordered intensional interpretation. Let \(\boldsymbol P^Q = (P^Q_1,\ldots,P^Q_{d_k}),\) \[
\boldsymbol P^K
=
(P^K_1,\ldots,P^K_{d_k}),
\]
and \(\boldsymbol P^V = (P^V_1,\ldots,P^V_{d_v})\) be the interpretations of the query, key, and value vectors respectively.

The first dynamically generated parameter structure of the attention head is
\[
K^\top
\in
\overline{\mathbb R}_{\wstar}^{\,d_k\times n}.
\]
When it parameterizes the transformation applied to a query vector
\[
\boldsymbol q_i\in\overline{\mathbb R}^{\,d_k},
\]
the resulting raw-attention vector belongs to \(\overline{\mathbb R}^{\,n}.\) Each column \(j\) of \(K^\top\) originates from the key vector \(\boldsymbol k_j\) and therefore constitutes a token-indexed point of view associated by provenance with token \(a_j\). Since the rows of \(K\) are interpretively uniform, all these columns share the same ordered second-level interpretation \((P^K_1,\ldots,P^K_{d_k})\) relative to the query interpretation \(\boldsymbol P^Q\).

At the level of the general framework, nothing prevents a second-level key interpretation from containing preserving properties \(\wwstar\). For the abstract attention-head type considered here, however, we impose the additional functional constraint that key positions are interpreted by ordinary properties:
\[
P^K_\ell\in\mathcal P,
\qquad
\ell=1,\ldots,d_k.
\]
This constraint is satisfied by the standard Transformer and excludes the limiting degenerate possibility in which every key position is interpreted by \(\wwstar\). In that case, the key structure would reproduce the query interpretation componentwise and would therefore fail to perform the semantic role characteristic of an attention head, as explained below.

Under the stated constraint,
\[
P^Q_\ell\odot P^K_\ell
=
P^K_\ell
\qquad
(\ell=1,\ldots,d_k).
\]
Hence every position of a raw attention vector is interpreted by the same property \(A := \bigoplus_{\ell=1}^{d_k}P^K_\ell.\) The property \(A\) is therefore characteristic of the attention head under consideration: it results from the combination of its key properties.

The semantically non-trivial case is that in which at least one key property differs from the corresponding query property, \(P^K_\ell\neq P^Q_\ell\) for some \(\ell\). In that case, at least one query property is replaced by projection with a more abstract key property, before the resulting key properties are combined into \(A\).

Property sum is abstraction-level preserving: the property resulting from the combination is neither less abstract than any of its component properties nor more abstract than each of them (see \S5.2). The abstraction involved in the construction of \(A\) is therefore not generated by property sum itself. It is already present in the key interpretation and is preserved when the key properties are combined. In the semantically non-trivial case, \(A\) is consequently more abstract than each query property.

However, the identity of \(A\) across the \(n\) positions does not imply identity of their values. Although all columns of \(K^\top\) share the same second-level intensional interpretation, their numerical values differ because each originates from the key vector associated with a different token \(a_j\). The implications associated with the key properties therefore depend not only on the properties involved but also, at the value level, on the particular point of view through which those properties are applied.

For a query originating from token \(a_i\), the \(j\)-th raw attention score is thus an occurrence-relative value of the same property \(A\), obtained from point of view \(j\) associated with token \(a_j\): the raw-attention vector has the uniform intensional interpretation \((A,\ldots,A),\) while its different positions generally have different values.

The subsequent rescaling and masking stages are intensionally invariant transforming concepts, while the normalization stage is an intensionally invariant operational block, instantiated by Softmax in the standard Transformer. They may modify the numerical values of the vector but do not introduce a different ordered intensional interpretation. The resulting attention vector
\[
\boldsymbol\alpha_i
=
(\alpha_{i1},\ldots,\alpha_{in})
\in
\overline{\mathbb R}^{\,n}
\]
therefore retains the interpretation \((A,\ldots,A).\) The first part of the attention pattern can thus be understood semantically as producing, for each token \(a_i\), different occurrence-relative values of one and the same more abstract attention property \(A\), each obtained through a different token-indexed point of view.

\subsection{Value aggregation and the semantic output of attention}

The second dynamically generated parameter structure involved in the attention pattern is
\[
V
\in
\overline{\mathbb R}_{\wstar}^{\,n\times d_v},
\]
when the value matrix assumes its parameter role. The attention vector
\[
\boldsymbol\alpha_i
\in
\overline{\mathbb R}^{\,n}
\]
is now the argument of a transformation whose result belongs to \(\overline{\mathbb R}^{\,d_v}.\) 
The rows \(\boldsymbol v_1,\ldots,\boldsymbol v_n\) of \(V\) share the common intensional interpretation \(\boldsymbol P^V = (P^V_1,\ldots,P^V_{d_v}),\) while retaining their distinct referential provenance from the corresponding tokens \(a_1,\ldots,a_n\).

For the abstract attention-head type considered here, we likewise require the second-level interpretation supplied by \(V\) to contain only ordinary properties: \(P^V_h\in\mathcal P, \qquad h=1,\ldots,d_v.\) This constraint, also satisfied by the standard Transformer, excludes preserving properties \(\wwstar\), which would preserve the attention property \(A\) instead of supplying the value property associated with the corresponding output position.

For output position \(h\), every argument position \(j\) is associated with the same second-level property \(P^V_h\). Since the attention vector is interpreted throughout by \(A\), the property-level result is
\[
R_h
:=
\bigoplus_{j=1}^{n}
\left(
A\odot P^V_h
\right).
\]
Under the stated constraint, \(A\odot P^V_h = P^V_h.\) By the idempotence of property sum, \(R_h = \bigoplus_{j=1}^{n}P^V_h = P^V_h.\) Hence the output vector
\[
\boldsymbol z_i
\in
\overline{\mathbb R}^{\,d_v}
\]
has the ordered intensional interpretation \(\boldsymbol P^V.\) It continues to refer to token \(a_i\), while its numerical values result from contributions whose provenance lies in the different token-indexed value vectors \(\boldsymbol v_j\).

This yields a general semantic explanation of the attention pattern. Its first part constructs a single attention property \(A\) from the key interpretation and assigns different values of that property according to token-indexed points of view. Its second part uses those values to regulate the contributions associated with the corresponding value vectors, while the resulting vector preserves the ordered intensional interpretation of the value representation and the reference of the query-originating token.

Together with the ordinary-property constraint just stated, two structural conditions are crucial to this result: \emph{interpretive uniformity} and the \emph{idempotence of property sum}. Because the value vectors share the same interpretation, each contribution to output position \(h\) implicates the same property \(P^V_h\). Because property sum is idempotent, combining any number of such contributions leaves that property unchanged.

The semantic explanation therefore does not depend on ordinary multiplication and addition as such, nor on the particular mathematical function used by the normalization operational block, such as Softmax in the standard Transformer. Different mathematical instantiations can alter the numerical values generated by the attention pattern while preserving the same semantic organization, provided that they instantiate the same local functional roles and satisfy the relevant interpretive constraints.

The resulting invariance is therefore not numerical but semantic and functional: mathematically different instantiations may realize the same organization of token-indexed points of view, attention values, regulated contributions, and intensional output.

\subsection{Residual connections and intensional change}

Residual connections exhibit a different semantic organization.

Let
\[
\boldsymbol x_i
=
(x_{i1},\ldots,x_{id})
\in
\overline{\mathbb R}^{\,d}
\]
be a representational vector entering an attention or feed-forward module. Suppose that it refers to token \(a_i\) and is interpreted by \(\boldsymbol P = (P_1,\ldots,P_d).\) Let
\[
\boldsymbol o_i
=
(o_{i1},\ldots,o_{id})
\in
\overline{\mathbb R}^{\,d}
\]
be the corresponding module output. By semantic coordination it refers, in its representational role, to the same token \(a_i\), but it may have a different ordered intensional interpretation \(\boldsymbol Q = (Q_1,\ldots,Q_d).\) 
In the abstract residual-connection type, \(\boldsymbol o_i\) becomes the dynamically generated parameter vector of the componentwise transformation
\[
C_{\boldsymbol o_i}^{\diamond^{\times}}(\boldsymbol x_i).
\]
In the standard Transformer, \(\diamond^{\times}\) is instantiated by \(+^{\times}\), yielding
\[
C_{\boldsymbol o_i}^{+^{\times}}(\boldsymbol x_i)
=
\boldsymbol x_i+\boldsymbol o_i.
\]

The referential provenance of \(\boldsymbol o_i\) remains architecturally relevant: it identifies the module output associated with the same token as the input vector \(\boldsymbol x_i\), thereby fixing the token-by-token correspondence required by the residual organization. Qua parameter vector, however, \(\boldsymbol o_i\) has no extensional reference of its own. Its properties acquire instead a second-level interpretation relative to the interpretation \(\boldsymbol P\) of the argument.

For the residual-connection type considered here, the second-level properties \(Q_j\) are ordinary properties. Hence \(P_j\odot Q_j = Q_j \qquad (j=1,\ldots,d).\) Consequently,
\[
C_{\boldsymbol Q_{\boldsymbol P}}^{\odot}
(\boldsymbol P)
=
\boldsymbol Q.
\]
The residual-connection output therefore has the intensional interpretation of the module output used as parameter. The residual transforming concept is intensionally invariant exactly when \(\boldsymbol Q=\boldsymbol P.\) When instead \(\boldsymbol Q\neq\boldsymbol P,\) the residual connection is dimension-preserving but intensionally changing.

This result clarifies a feature that the value-level expression
\[
C_{\boldsymbol o_i}^{\diamond^{\times}}(\boldsymbol x_i)
\]
does not by itself reveal. The two vectors have the same dimensionality, belong to the same numerical domain, and, in their representational roles, refer to the same token. Yet they need not have the same intensional interpretation. Their mathematical similarity therefore does not entail semantic identity.

Moreover, the two vectors occupy different semantic roles. The input vector is the argument, whereas the module output becomes the dynamically generated parameter structure. This asymmetry is fixed by the output--parameter interconnection independently of the particular numerical operation instantiating \(\diamond^{\times}\). In the standard Transformer this point is especially striking because \(\diamond^{\times}\) is instantiated by \(+^{\times}\), where \(+\), i.e. ordinary addition, is symmetric, although the semantic organization of its two operands is not.

Residual connections therefore provide a concrete instance of the general result established in \S5.4: \emph{dimensional preservation does not imply intensional invariance}.

\subsection{Complementarity and modularity}

The two analyses reveal a complementary pattern, although this is not their principal result. In attention, different mathematical operations and functions can preserve semantic organization through interpretive uniformity, referential provenance, projection, and idempotent property sum. Residual connections show the converse possibility: vectors of the same mathematical kind may play different semantic roles and carry different intensional interpretations, so mathematical similarity does not determine their semantic relations.

More importantly, an attention head and a residual connection can themselves be regarded as different abstract types of semantically coordinated systems, each with a local functional architecture governed by the common algebraic-semantic core. This modular organization is already implicit in Figure~1 of TR1; the present framework makes its significance explicit. A complex architecture can therefore be decomposed into simpler modular types that may recur, with the same semantic analysis, across otherwise different abstract architectures. The framework thus reveals semantic invariance across mathematical variation, semantic differences despite mathematical similarity, and modular semantic organization across system types.

\section{Philosophical Significance, Scope, and Conclusion}

The preceding results identify a level at which the semantic organization of artificial systems can be characterized without reducing it to numerical description or presupposing a theory of representational grounding. Their philosophical significance concerns two connected issues: the relation between \emph{formal semantics, representation, and metasemantic grounding}, and \emph{sameness and variation across descriptions and instantiations}. The modular analyses of attention and residual connections extend these questions from complete architectures to functional subsystems.

\subsection{Formal semantics before representation and grounding}

Contemporary debates about representation and content in artificial systems commonly ask under what conditions internal states genuinely represent objects, properties, or concepts. The approaches considered in \S2 emphasize different candidates for answering this question: coactivation and functional role, informational relations, downstream use, conditions for misrepresentation, structural correspondence, or more general theories of aboutness and metasemantics (Freiesleben 2026; Poth and Schuster 2026; Klein 2026; Milli\`ere and Buckner 2026; Cappelen and Dever 2021). Kieval's (2026) warning against inferring philosophical representationalism directly from the success of representation learning is especially relevant here.

The present proposal does not attempt to decide among these accounts. It changes the order of analysis. Before asking what makes a vector, a position, or a property genuinely representational, one can ask what formal semantic organization is obtained once the relevant interpretations are specified.

C1 provides such an organization by distinguishing extensional vector reference, ordered intensional interpretation, and occurrence-relative property extension, while parameter structures receive intensional interpretations without object-reference. It thereby separates formal interpretation, genuine representational relations, and their grounding.

The priority claimed is \emph{analytical rather than metasemantic}. Formal interpretation neither creates content nor explains why a given property or referent is correct; it specifies what a theory of representation or grounding must explain, including the assignment of properties to positions and of referents to representational vectors.

This gives formal semantics an autonomous philosophical role without turning it into a substitute for metasemantics. By articulating the structure of interpretation before committing to a particular account of its grounding, the framework also follows a methodological orientation naturally associated with the formal-semantic tradition represented by Dalla Chiara's work: formal distinctions concerning interpretation can themselves be philosophically informative while leaving stronger questions concerning their foundation open (Dalla Chiara et al. 2018).

\subsection{From representing contents to transforming them}

The second contribution goes beyond the formal semantics of representational states. C2 extends semantic analysis to the transformations acting on them.

This extension matters because the semantic organization of the systems considered here is not exhausted by assigning contents to vectors. Parameter structures acquire relative, second-level interpretations with respect to the properties interpreting their arguments. A parameterized transformation becomes a \emph{transforming concept} only when its numerical action is coordinated with the corresponding property-level map. Mathematical parameterization and conceptual parameterization are therefore distinct.

If internal representations are transformed during processing, it is not enough to ask what successive vectors represent: one must also characterize how one interpreted representation becomes another and the semantic role of the structures determining that change. Projection, property sum, relative interpretations, and interpretive points of view make this explicit. Parameter structures can therefore play a relational semantic role without representing objects themselves, specifying how argument properties are preserved, replaced, or combined; in output--parameter interconnections this role may arise dynamically without erasing referential provenance.

The distinction between constitutive semantics and genesis becomes especially important here. Whether a transforming concept is static or dynamic, whether its parameter values arise through training or during inference, and whether its interpretation depends on environmental interaction or architectural determination do not define its status as a transforming concept. Those questions concern how the relevant semantic organization comes about; the formal semantics specifies what that organization is once present.

The proposal therefore extends the philosophical problem of content in artificial systems from \emph{content-bearing states to content-transforming organization}. This is precisely what makes the analyses in \S8 possible: attention and residual connections can be explained semantically not merely by interpreting their inputs and outputs, but by reconstructing the semantic roles performed by the transformations and interconnections that relate them.

\subsection{Sameness, variation, and modular explanation}

C3 introduces a second form of abstraction. Different mathematical systems can instantiate the same abstract functional architecture when they preserve both the common algebraic-semantic core and the architecture-specific organization of transformations, operations, interconnections, and functional roles. Their identity as systems of the relevant abstract type therefore does not require identity of their numerical operations or of the mathematical functions assigned to particular operational blocks.

This notion of \emph{functional-architectural invariance} is related to, but should not be conflated with, familiar discussions of multiple realization. The latter typically concern sameness at one level together with physically or mechanically relevant differences at another (Polger and Shapiro 2016). Recent work has renewed this question in connection with deep learning and has emphasized that multiple realizability does not make differences of implementation explanatorily irrelevant (McGrath and Russin 2024). The invariance established here occurs at a different boundary: it concerns different \emph{mathematical instantiations of the same abstract functional type}, prior to asking how that type is mechanically or physically realized.

This distinction makes the relations among levels more articulated than a single hierarchy from abstract description to physical realization. As shown in \S7.5, a mathematically instantiated abstract functional system can be computationally implemented through computational simulation without first being mechanically realized. Alternatively, the abstract system may receive a mechanical realization, which can receive the same mathematical instantiation and subsequently guide a physical construction. Computational implementation and physical construction are therefore different routes to the physical level rather than successive stages of one necessary route.

Section~8 adds \emph{modularity of semantic explanation}: attention heads and residual connections are themselves abstract types whose explanation depends on local functional organization and the common core, so the same module may recur across different larger architectures. Attention also shows semantic organization preserved across mathematical instantiations, while residual connections show that vectors of the same mathematical kind may have different semantic roles and intensional interpretations. The relevant relations are fixed by coordination between the algebraic-semantic core and functional organization, not by mathematical similarity alone.

\subsection{Scope and open questions}

These results also delimit what the present framework establishes.

First, the framework is \emph{not a general theory of representation}: it presupposes that some vectors represent objects and does not derive that status or their properties from mathematical structure. Second, it is \emph{not a metasemantic theory}: causal/informational, functional/inferential, environmental, historical, or other determinants of content may explain why the formal interpretations are appropriate, and first- and second-level interpretation, extensional reference, occurrence-relative extension, and referential provenance need not all receive the same metasemantic explanation. Third, the genetic classification of \S5.5 is preliminary and does not explain how systems acquire content.

Finally, the paper does not theorize the mechanical, neural, or physical realization of these abstract systems. Section~7 only distinguishes mathematical instantiation, mechanical realization, computational implementation, and physical construction. The further questions of generalized neural realizability and of how the present generalization bears on Conjecture T---the in-principle hypothesis introduced in TR1 that human neural systems may realize a functional architecture relevantly similar to that of the Transformer---are left for future work. Taken together, these limitations preserve distinct explanatory levels so that generation, grounding, realization, implementation, and construction can later be investigated separately.

\subsection{Conclusion}

The starting point of this paper was the interpretation proposed in TR1 of the Transformer as a system that transforms concepts by means of concepts. The aim has been to determine whether the semantic organization suggested by that interpretation can be formulated independently of the particular mathematics and neural organization of the standard Transformer and generalized to a wider class of systems.

The framework yields three main results: \textbf{C1} distinguishes the intensional--extensional semantics of representational vectors from the non-referential interpretation of parameter structures; \textbf{C2} defines transforming concepts by coordinating value-level transformations with property-level maps determined by relative, second-level interpretations; and \textbf{C3} separates abstract functional architecture from mathematical instantiation and mechanical realization, establishing functional-architectural invariance and two routes to the physical level.

Section~8 shows the explanatory payoff: attention and residual semantics can remain invariant across different mathematical instantiations of the same local organization, and both patterns are modular abstract system types. The resulting generalization is therefore both \emph{mathematical and architectural}.

The framework does not show that every vector transformation is semantic, that every mathematically adequate model represents, or that formal interpretation is sufficient for genuine content. It establishes a more limited but prior result: \emph{if coordinated vector transformations are semantically interpreted in the manner specified here, their semantic organization can itself be formally characterized, their transformations can be analyzed as transformations of interpreted properties, and the invariants of that organization can be separated from the particular mathematical, mechanical, and physical forms through which it is instantiated or realized.}

This provides a basis for moving to the next question rather than an answer to it: how abstract systems with this semantic and functional organization can be mechanically---and, in particular, neurally---realized, computationally implemented, or physically constructed.

\section*{References}
\begingroup
\small
\setlength{\parindent}{0pt}
\setlength{\parskip}{0.65em}

Giunti, Marco, and Fabrizia Giulia Garavaglia. 2026. ``The Transformer Revolution, Part 1: Dynamic Processing through Output-Weight Interconnections.'' arXiv. \url{https://doi.org/10.48550/arXiv.2608.03921}.

Buckner, Cameron. 2018. ``Empiricism without Magic: Transformational Abstraction in Deep Convolutional Neural Networks.'' \emph{Synthese} 195 (12): 5339--5372. \url{https://doi.org/10.1007/s11229-018-01949-1}.

Cappelen, Herman, and Josh Dever. 2021. \emph{Making AI Intelligible: Philosophical Foundations}. Oxford: Oxford University Press. \url{https://doi.org/10.1093/oso/9780192894724.001.0001}.

Dalla Chiara, Maria Luisa, Roberto Giuntini, Roberto Leporini, and Giuseppe Sergioli. 2018. \emph{Quantum Computation and Logic: How Quantum Computers Have Inspired Logical Investigations}. Cham: Springer. \url{https://doi.org/10.1007/978-3-030-04471-8}.

Freiesleben, Timo. 2026. ``Artificial Neural Nets and the Representation of Human Concepts.'' In \emph{Philosophy of Science for Machine Learning: Core Issues and New Perspectives}, edited by Juan M. Dur\'an and Giorgia Pozzi, 273--294. Cham: Springer. \url{https://doi.org/10.1007/978-3-032-03083-2_13}.

Humphreys, Paul. 1990. ``Computer Simulations.'' \emph{PSA: Proceedings of the Biennial Meeting of the Philosophy of Science Association} 1990 (2): 497--506.

Kieval, Phillip Hintikka. 2026. ``Representation Learning Without Representationalism: A Non-representational Account of Deep Learning Models in Scientific Practice.'' In \emph{Philosophy of Science for Machine Learning: Core Issues and New Perspectives}, edited by Juan M. Dur\'an and Giorgia Pozzi, 245--272. Cham: Springer. \url{https://doi.org/10.1007/978-3-032-03083-2_12}.

Klein, Colin. 2026. ``Represented Structure versus Representational Structure: A Challenge for Interpreting LLMs.'' \emph{Philosophy and the Mind Sciences} 7 (1): 1--12. \url{https://doi.org/10.33735/phimisci.2026.12137}.

McGrath, Sam Whitman, and Jacob Russin. 2024. ``Multiple Realizability and the Rise of Deep Learning.'' arXiv:2405.13231. \url{https://doi.org/10.48550/arXiv.2405.13231}.

Milli\`ere, Rapha\"el, and Cameron Buckner. 2026. ``The Philosophy of Language Models.'' \emph{Philosophy Compass} 21 (3): e70095. \url{https://doi.org/10.1111/phc3.70095}.

Polger, Thomas W., and Lawrence A. Shapiro. 2016. \emph{The Multiple Realization Book}. Oxford: Oxford University Press. \url{https://doi.org/10.1093/acprof:oso/9780198732891.001.0001}.

Poth, Nina, and Annika Schuster. 2026. ``Mental, Scientific, and Artificial Representations.'' \emph{Philosophy and the Mind Sciences} 7 (1). \url{https://doi.org/10.33735/phimisci.2026.12094}.

R\"az, Tim. 2023. ``Methods for Identifying Emergent Concepts in Deep Neural Networks.'' \emph{Patterns} 4 (6): 100761. \url{https://doi.org/10.1016/j.patter.2023.100761}.

Vaswani, Ashish, Noam Shazeer, Niki Parmar, Jakob Uszkoreit, Llion Jones, Aidan N. Gomez, \L{}ukasz Kaiser, and Illia Polosukhin. 2017. ``Attention Is All You Need.'' In \emph{Advances in Neural Information Processing Systems 30}, 5998--6008.

\endgroup

\end{document}